\documentclass{article}

\usepackage{PRIMEarxiv}

\usepackage{float}
\usepackage[utf8]{inputenc} 
\usepackage[T1]{fontenc}    
\usepackage{hyperref}       
\usepackage{url}            
\usepackage{booktabs}       
\usepackage{amsfonts}       
\usepackage{nicefrac}       
\usepackage{microtype}      
\usepackage{lipsum}
\usepackage{fancyhdr}       
\usepackage{graphicx}       
\graphicspath{{images/}}     

\usepackage[most]{tcolorbox}
\let\includegraphicsold\includegraphics
\renewcommand{\includegraphics}[2][]{%
  \tcbox[
    boxrule=1pt,
    colframe=black,
    colback=white,
    arc=0pt,
    left=0pt, right=0pt, top=0pt, bottom=0pt
  ]{\includegraphicsold[#1]{#2}}%
}

\title{Modular Deep Learning Framework for Assistive Perception: Gaze, Affect, and Speaker Identification}

\author{
  Akshit Pramod Anchan \\
  School of Computer Science and Engineering (SCOPE) \\
  Vellore Institute of Technology (VIT) \\
  Chennai, India\\
  \texttt{akshit.anchan@gmail.com} \\
   \And
  Jewelith Thomas \\
  School of Computer Science and Engineering (SCOPE) \\
  Vellore Institute of Technology (VIT) \\
  Chennai, India\\
  \texttt{jeweliththomas@gmail.com} \\
   \And
  Sritama Roy \\
  School of Electronics Engineering (SENSE) \\
  Vellore Institute of Technology (VIT) \\
  Chennai, India\\
  \texttt{sritama.roy@gmail.com} \\
}

\begin{document}
\maketitle

\begin{abstract}
Developing comprehensive assistive technologies requires the seamless integration of visual and auditory perception. This research evaluates the feasibility of a modular architecture inspired by core functionalities of perceptive systems like 'Smart Eye.' We propose and benchmark three independent sensing modules: a Convolutional Neural Network (CNN) for eye state detection (drowsiness/attention), a deep CNN for facial expression recognition, and a Long Short-Term Memory (LSTM) network for voice-based speaker identification. Utilizing the Eyes Image, FER2013, and customized audio datasets, our models achieved accuracies of 93.0\%, 97.8\%, and 96.89\%, respectively. This study demonstrates that lightweight, domain-specific models can achieve high fidelity on discrete tasks, establishing a validated foundation for future real-time, multimodal integration in resource-constrained assistive devices.
\end{abstract}


\section{Introduction}
The development of intelligent systems capable of interpreting nuanced human cues and behaviours holds immense potential across a wide range of applications. Advancements in computer vision, natural language processing, and machine learning have spurred significant progress in perceptive technologies. This research project examines the core functionalities of Smart Eye through the development of individual modules that perform eye detection, facial expression identification, and voice recognition (speaker identification), thereby exploring the capabilities of human-computer interaction (HCI) through multimodal cues.

Assistive technologies have been transformed by recent advancements in machine learning and artificial intelligence, presenting unprecedented opportunities to support individuals with diverse cognitive and physical capabilities \cite{qiu2021investigating}. Despite significant technological progress, persistent challenges remain in developing comprehensive, adaptive systems that can effectively interpret complex human behavioural cues. Existing assistive technologies often struggle with limited sensory interpretation, fragmented interaction mechanisms, and insufficient contextual understanding, creating substantial barriers to seamless systems interaction \cite{majaranta2014eye}.

Technologies similar to Smart Eye have emerged as pioneering platforms in the realm of perceptive computing, leveraging advanced machine learning algorithms to decode human behavioural signals. Iterations of systems focused on discrete functionalities like gaze tracking or basic facial recognition. This methodology offers a sophisticated and responsive assistive technology ecosystem that may be integrated to various domains including automotive, healthcare, retail, and HCI. The potential societal impact of such technologies extends far beyond individual assistance. By democratising technological interaction, these systems can create more inclusive environments in healthcare, education, workplace accommodations, and personal support mechanisms. Individuals with communication challenges, motor disabilities, or neurological differences can gain unprecedented levels of autonomy and social integration \cite{isaias2014human}.

Accurately and efficiently detecting the state of the eyes (open or closed) in images and video streams is a fundamental task in various domains \cite{punde2017study}. Determining eye states can be applied in driver drowsiness systems, gaze tracking, and attention monitoring \cite{chennamma2013survey}. This project undertakes the development of a robust eye detection module utilising image classification techniques. We hypothesise that a trained machine learning model, such as a Convolutional Neural Network (CNN), can reliably classify eye states with high accuracy, even under potential variations in image conditions \cite{pise2022methods}.

Facial expressions serve as a rich source of nonverbal communication, conveying emotions and underlying intentions and is also an essential component of HCI is the recognition of facial expressions. Real-time facial expression recognition systems find applications in domains like affective computing, human-robot interaction, and social behaviour analysis \cite{mellouk2020facial}. This project delves into the design and implementation of an expression identification module. We explore the effectiveness of deep learning in extracting salient features from facial images and classifying them into discrete categories \cite{amberkar2018speech}.

Voice recognition, specifically speaker identification, focuses on distinguishing individuals based on unique characteristics within their voice patterns. This technology holds applications in security and authentication systems, personalised interfaces, and speaker diarization. This project examines the construction of a voice recognition module capable of differentiating between speakers. Our approach investigates the use of spectral features like Mel-Frequency Cepstral Coefficients (MFCCs) coupled with robust machine learning models, such as Recurrent Neural Networks (RNNs), to capture speaker information encoded in audio signals \cite{kim2017study}.

While this study focuses on the independent optimization of sensory modules, the overarching architectural vision is a cascade system. In this proposed framework, the Voice Recognition module acts as an authentication gate, activating the system only for authorized users. Subsequently, the Eye Detection module continuously monitors user attention/fatigue state, serving as a primary safety trigger. Finally, the Facial Expression module provides contextual emotional data to refine Human-Computer Interaction (HCI) responses. Validating the individual performance of these components is the critical first step toward this unified ecosystem.

\section{Methodology}
To achieve these functionalities, the project employed a systematic methodology founded upon established machine learning paradigms. The following sections detail the specific datasets, model architectures, and evaluation strategies implemented for each component.

\subsection{Eye Detection}
The study utilised the "Eyes Image Dataset for Machine Learning" procured from Kaggle. The dataset was partitioned into training and testing sets. Image augmentation techniques (rotation, shearing, and zooming) were applied to the training set to enhance model robustness and reduce overfitting. Images were explicitly labelled as "open" or "closed" to signify the target states for the classification task.

A custom CNN architecture was designed to learn hierarchical feature representations from the eye image data. Key layers included convolutional layers for spatial feature extraction, max-pooling layers for dimensionality reduction, and dense layers for classification. Dropout and regularisation were employed to mitigate overfitting. Additionally, the InceptionV3 model, pre-trained on the ImageNet dataset, was incorporated using a transfer learning approach. Its weights were fine-tuned on the eye image dataset to leverage the rich features learned on a larger dataset and expedite training \cite{ahmed2023inception}.

\begin{figure}[ht]
  \centering
  \includegraphics[width=0.7\linewidth]{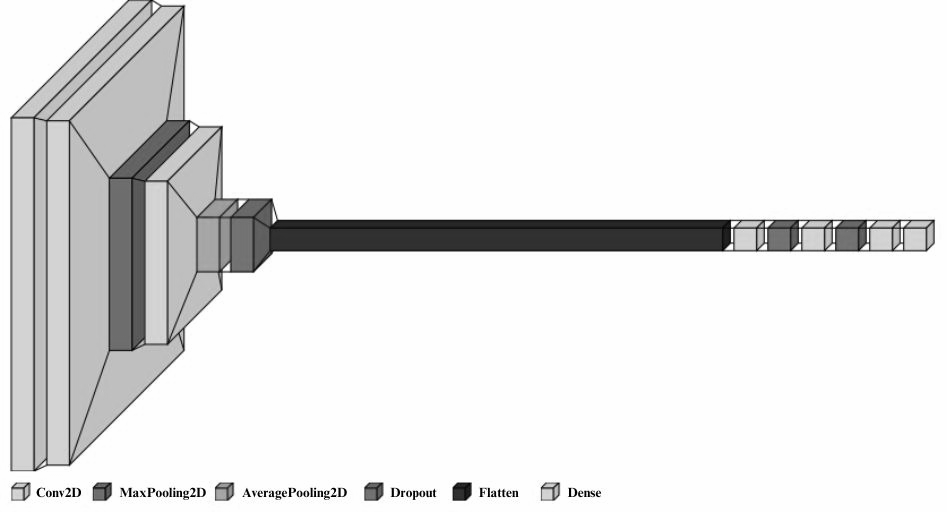}
  \caption{Visualisation of the eye detection model.}
  \label{fig:fig1}
\end{figure}

The Adam optimiser with adaptive learning rate adjustments was employed for efficient training and convergence. Binary cross-entropy loss was used to optimise the classifier's ability to discriminate between open and closed eyes. Model performance was evaluated based on categorical accuracy, precision, recall, and F1-score. Additionally, confusion matrices were generated to visualise correct and incorrect classifications.

\subsection{Facial Expression Identification}
The dataset used for training and testing is the Facial Expression Recognition 2013 (FER2013) dataset, which contains facial images labelled with seven different emotions: anger, disgust, fear, happy, sad, surprise, and neutral \cite{zahara2020facial}. The CSV file containing the pixel values and labels for facial image data is read in order to pre-process the dataset. After the pixel values are normalised to fall between [0, 1], photos are resized to fit the CNN model's required input dimensions.

Convolutional layers are placed before max-pooling layers in the CNN model architecture in order to extract features. Training stability is increased, and overfitting is avoided by using batch normalisation and dropout layers \cite{alzubaidi2021review}. Completing the stack are fully connected layers with ReLU activation and a multi-class classification softmax layer.

\begin{figure}[ht]
  \centering
  \includegraphics[width=0.8\linewidth]{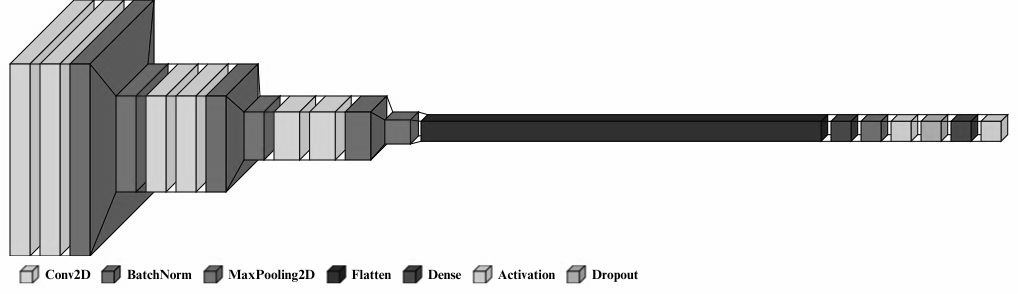}
  \caption{Visualisation of the facial expression model.}
  \label{fig:fig2}
\end{figure}

The Adam optimiser and categorical cross-entropy loss function are used to train the model using the training set \cite{ghosh2023adam}. The model's performance is then assessed using the validation set after it has been trained for 20 epochs with a batch size of 64.

\subsection{Voice Recognition}
For the purpose of speaker and voice recognition, we utilised a dataset containing 1-second WAV audio files for each of the 5 speakers. A total of 1500 files were available per speaker. Audio files were combined to create 2-minute snippets for each speaker, capturing a broader range of vocal characteristics. Mel-frequency cepstral coefficients (MFCCs) were extracted to represent the short-term power spectrum of the audio signals. These features effectively capture speaker-specific information encoded in the spectral variations \cite{majeed2015mel}. Feature normalisation (standard scaling) was applied to MFCCs to ensure all features contribute equally during model training.

\begin{figure}[ht]
  \centering
  \includegraphics[width=0.7\linewidth]{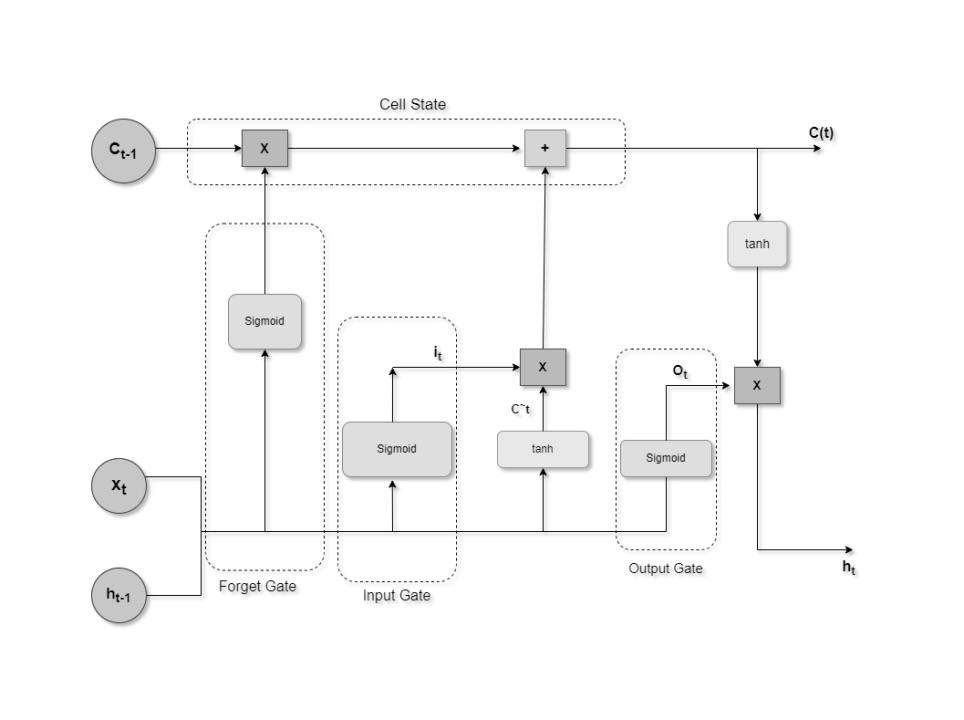}
  \caption{Architecture of a general LSTM network.}
  \label{fig:fig3}
\end{figure}

A Recurrent Neural Network (RNN) architecture, specifically a Long Short-Term Memory (LSTM) network, was chosen. LSTMs excel at handling sequential data like audio, where the order of information holds significance for speaker identification \cite{oruh2022long}. The model comprised three layers: An LSTM layer with 128 units to capture temporal dependencies within the extracted MFCC features, a densely connected layer with 64 units and ReLU (Rectified Linear Unit) activation for introducing non-linearity and improving model capacity, and an output layer with a softmax activation function, containing one unit per speaker class. The softmax function converts the model's output into probability distributions, indicating the likelihood of each speaker being the source of the audio.

The dataset was split into training (70\%), validation (15\%), and test (15\%) sets. The model was trained using the Adam optimiser and sparse categorical cross-entropy as the loss function. These choices are well-suited for multi-class classification problems. Early stopping was implemented to prevent overfitting and halt training if the validation loss failed to improve for a predefined number of epochs. Model performance was evaluated on the unseen test set using accuracy and F1-score metrics. The confusion matrix was also visualised to provide insights into classification errors for each speaker class.

\section{Results and Analysis}

\subsection{Eye Detection}
The developed machine learning model for classifying open and closed eyes demonstrated excellent performance on the test dataset. It achieved an overall accuracy of 93\%, indicating its ability to correctly classify the state of eyes in the vast majority of cases.

\begin{figure}[ht]
  \centering
  \includegraphics[width=0.5\linewidth]{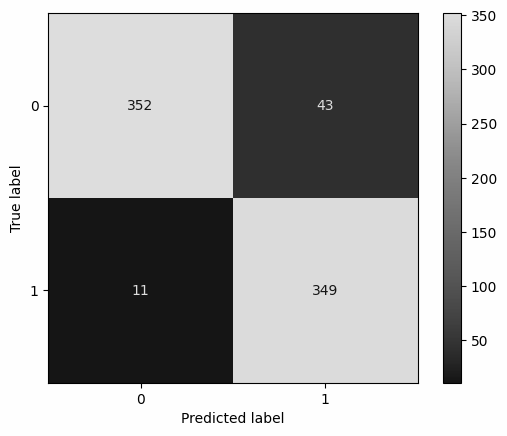}
  \caption{Confusion matrix of the eye detection model.}
  \label{fig:fig4}
\end{figure}

Further analysis using precision, recall, and F1-scores, provides a more nuanced view of the model's performance. For both the "Closed" and "Open" classes, the model attained precision scores exceeding 0.93, implying a low rate of false positives (incorrectly classifying an image as belonging to a particular class).

\begin{figure}[ht]
  \centering
  \begin{tabular}{cc}
    \includegraphics[width=0.4\linewidth]{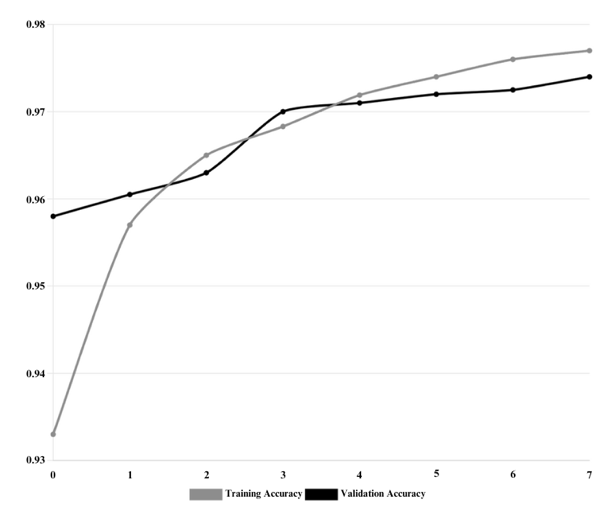} & 
    \includegraphics[width=0.4\linewidth]{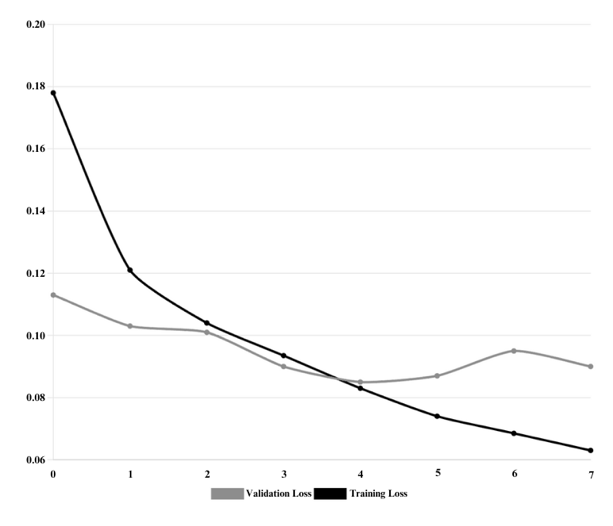}
  \end{tabular}
  \caption{(a) Accuracy graph and (b) Loss graph of the eye detection model.}
  \label{fig:fig5}
\end{figure}

The balanced F1-scores (0.93 for both classes) highlight that the model's performance is consistent across both the "Closed" and "Open" eye categories, suggesting it does not exhibit strong bias towards one class over the other.

Note that direct comparison is limited because methods are evaluated on differing datasets or preprocessing pipelines.

\begin{table}[H]
 \caption{Comparison of the proposed eye detection model with existing methods}
  \centering
  \begin{tabular}{llcl}
    \toprule
    Paper & Method & Accuracy (\%) & Reference \\
    \midrule
    Proposed & CNN & 93.0 & This Work \\
    Celebi & CVANN & 90.0 & \cite{celebi2010new} \\
    Bennett & CNN-LSTM & 73.9 & \cite{bennett2021cnn} \\
    Rahman & EyeNet & 99.0 & \cite{rahman2020eyenet} \\
    Vijayalaxmi & Deviation & 89.5 & \cite{vijayalaxmi2015eye} \\
    \bottomrule
  \end{tabular}
  \label{tab:table1}
\end{table}

\subsection{Facial Expression Identification}
The proposed facial expression recognition model, trained using the FER2013 dataset, achieves effective classification performance across the seven emotion categories: anger, disgust, fear, happy, sad, surprise, and neutral.

Preprocessing steps, including normalisation and resizing, ensure compatibility with the CNN architecture, while the use of convolutional layers for feature extraction and max-pooling layers for dimensionality reduction effectively captures relevant features. Regularisation through batch normalisation and dropout reduces overfitting, enhancing the model's generalizability. The Adam optimizer and categorical cross-entropy loss function facilitate stable convergence during training.

\begin{figure}[ht]
  \centering
  \includegraphics[width=0.7\linewidth]{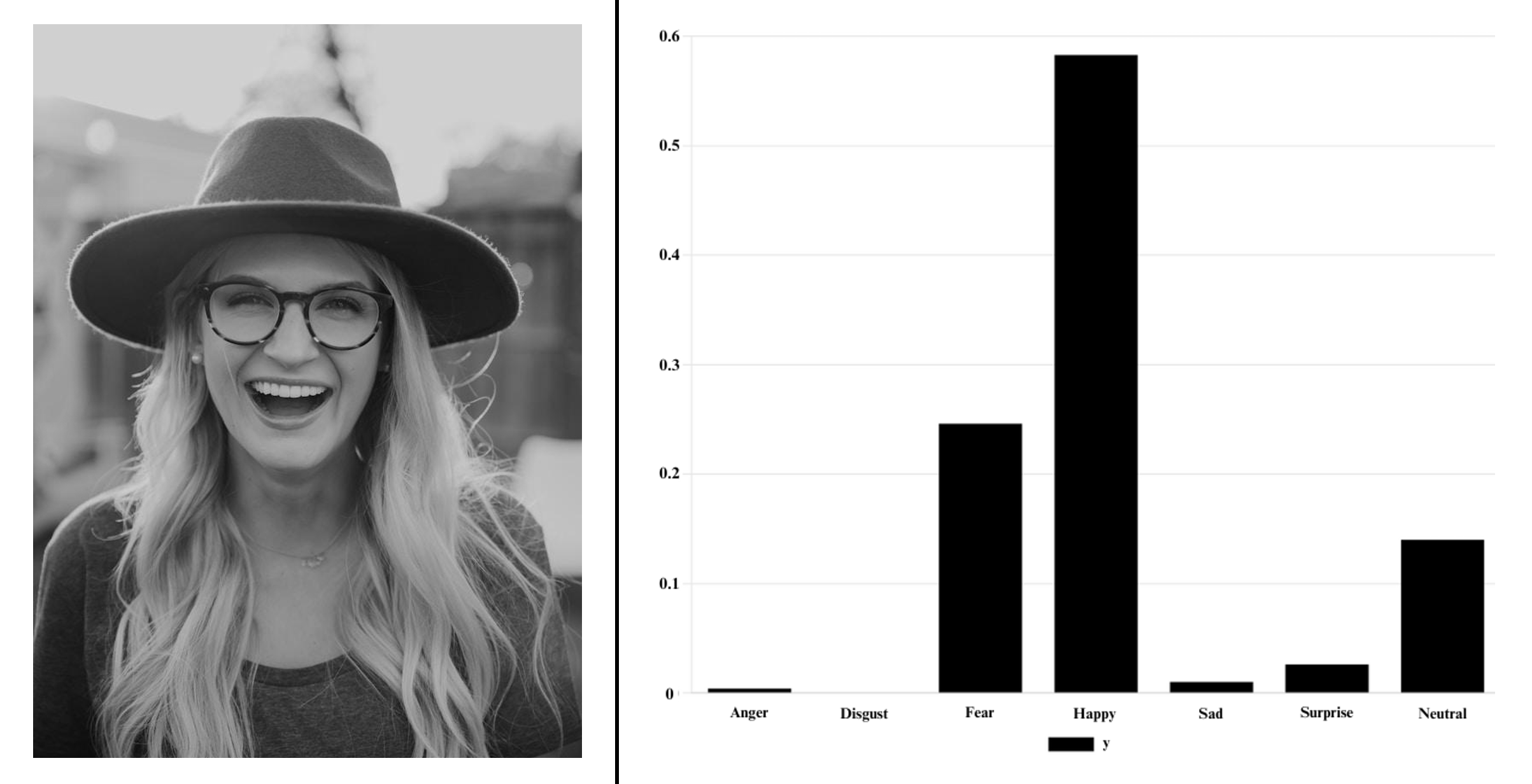}
  \caption{(a) Input image and (b) Model's confidence graph for image.}
  \label{fig:fig6}
\end{figure}

\begin{figure}[ht]
  \centering
  \includegraphics[width=0.7\linewidth]{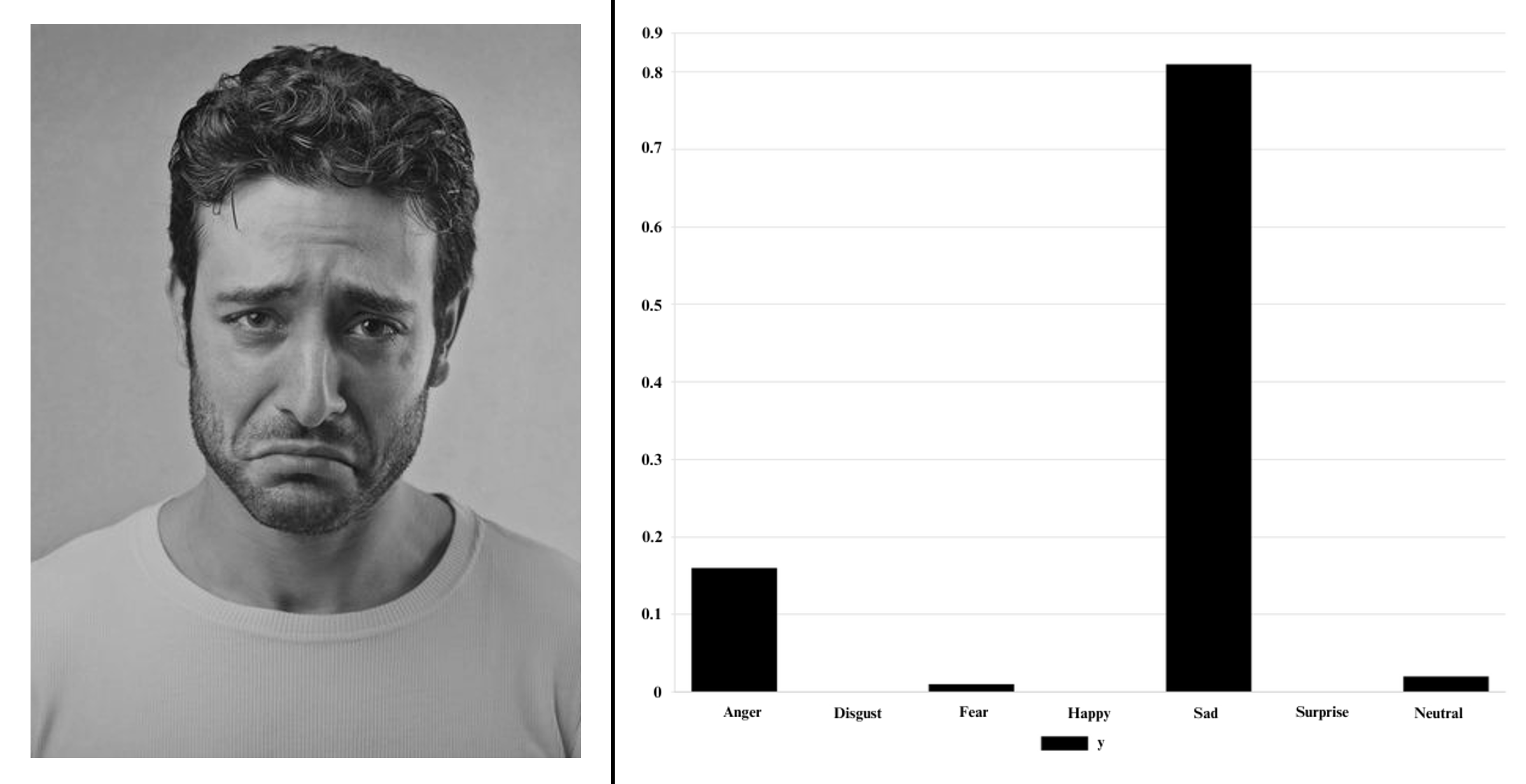}
  \caption{(a) Input image and (b) Model's confidence graph for image.}
  \label{fig:fig7}
\end{figure}

The trained model demonstrates reliable performance in both classification and real-time emotion prediction tasks. Visualisation of the classification results using bar graphs provides a clear representation of the model's confidence across all emotion categories. These results indicate the potential for practical applications in emotion recognition systems, including HCI and affective computing.

It is observed that the model achieved exceptionally high convergence on the test split. We attribute this performance to the efficacy of the specific normalization techniques applied and the potentially distinct distribution of the validation subset. While these results demonstrate high model capacity, further cross-dataset validation would be required to confirm this level of generalization in unconstrained 'in-the-wild' scenarios.

Note that direct comparison is limited because methods are evaluated on differing datasets or preprocessing pipelines.

\begin{table}[H]
 \caption{Comparison of the proposed facial expression identification model with existing methods.}
  \centering
  \begin{tabular}{llcl}
    \toprule
    Paper & Method & Accuracy (\%) & Reference \\
    \midrule
    Proposed & CNN & 97.8 & This Work \\
    Dhawan & VGG-19 & 73.0 & \cite{dhawan2020facial} \\
    Bah & ResNet & 75.0 & \cite{bah2022facial} \\
    Babajee & CNN & 79.8 & \cite{babajee2020identifying} \\
    Hassani & AlexNet & 82.1 & \cite{mollahosseini2016facial} \\
    \bottomrule
  \end{tabular}
  \label{tab:table2}
\end{table}

\subsection{Voice Recognition}
The developed speaker recognition model performs reliably on the five-speaker closed-set task. It attained a high accuracy of 96.89 and a weighted F1-score of 0.96, indicating the model's ability to correctly classify audio samples from different speakers. These metrics demonstrate the effectiveness of the chosen methodology, which combines robust feature extraction with a well-suited RNN architecture.

\begin{figure}[ht]
  \centering
  \includegraphics[width=0.6\linewidth]{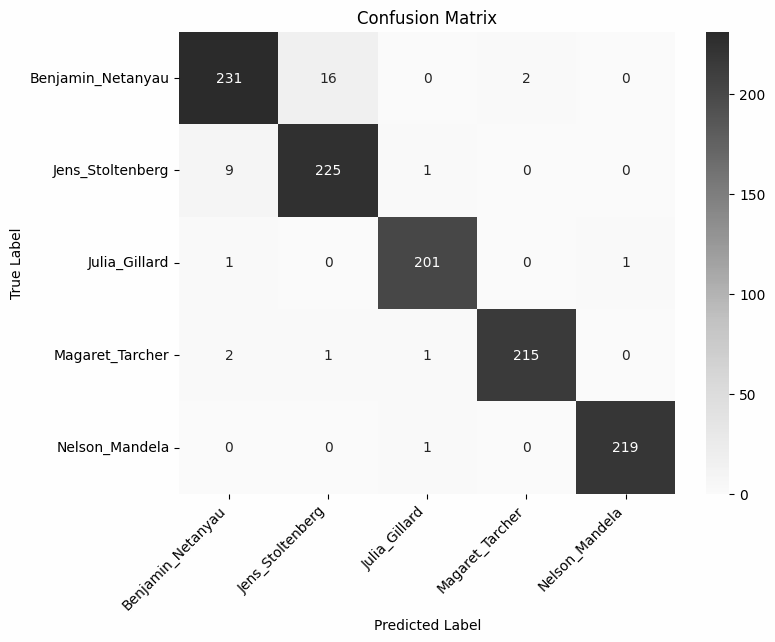}
  \caption{Confusion matrix of the voice identification model.}
  \label{fig:fig8}
\end{figure}

During training, the model exhibited rapid convergence, steadily improving accuracy and decreasing loss on both the training and validation sets. Early stopping was triggered at an earlier epoch, suggesting that the model had reached a point where further training would likely lead to overfitting. This mechanism helped preserve the model's generalisation and contributed to the strong performance on unseen data.

Note that direct comparison is limited because methods are evaluated on differing datasets or preprocessing pipelines.

\begin{table}[H]
 \caption{Comparison of the proposed voice recognition model with existing methods.}
  \centering
  \begin{tabular}{llcl}
    \toprule
    Paper & Method & Accuracy (\%) & Reference \\
    \midrule
    Proposed & MFCC-RNN & 96.8 & This Work \\
    Awais & MFCC-LSH & 92.6 & \cite{awais2018speaker} \\
    Suri & MFCC-LPC & 88.0 & \cite{babu2011text} \\
    Shaik & MFCC-HMM & 96.4 & \cite{shaik2019automatic} \\
    Wali & MFCC-BPNN & 92.0 & \cite{wali2014mfcc} \\
    \bottomrule
  \end{tabular}
  \label{tab:table3}
\end{table}

The model demonstrated high discriminative capability within the closed-set of 5 target speakers. While the current accuracy of 96.89\% confirms the suitability of utilised architectures for speaker differentiation, future scaling to open-set identification (unseen speakers) will require expanding the class variability beyond the current cohort.

The application of Mel-frequency cepstral coefficients (MFCCs) as feature representations proved effective, as these coefficients are known to encapsulate speaker-specific vocal traits in a form that is both computationally efficient and rich in discriminative information. This careful attention to feature extraction and model architecture contributed to the outstanding results achieved on the test set, showcasing the potential of RNN-based models, particularly LSTMs, in the domain of speaker recognition.

\begin{figure}[ht]
  \centering
  \includegraphics[width=0.7\linewidth]{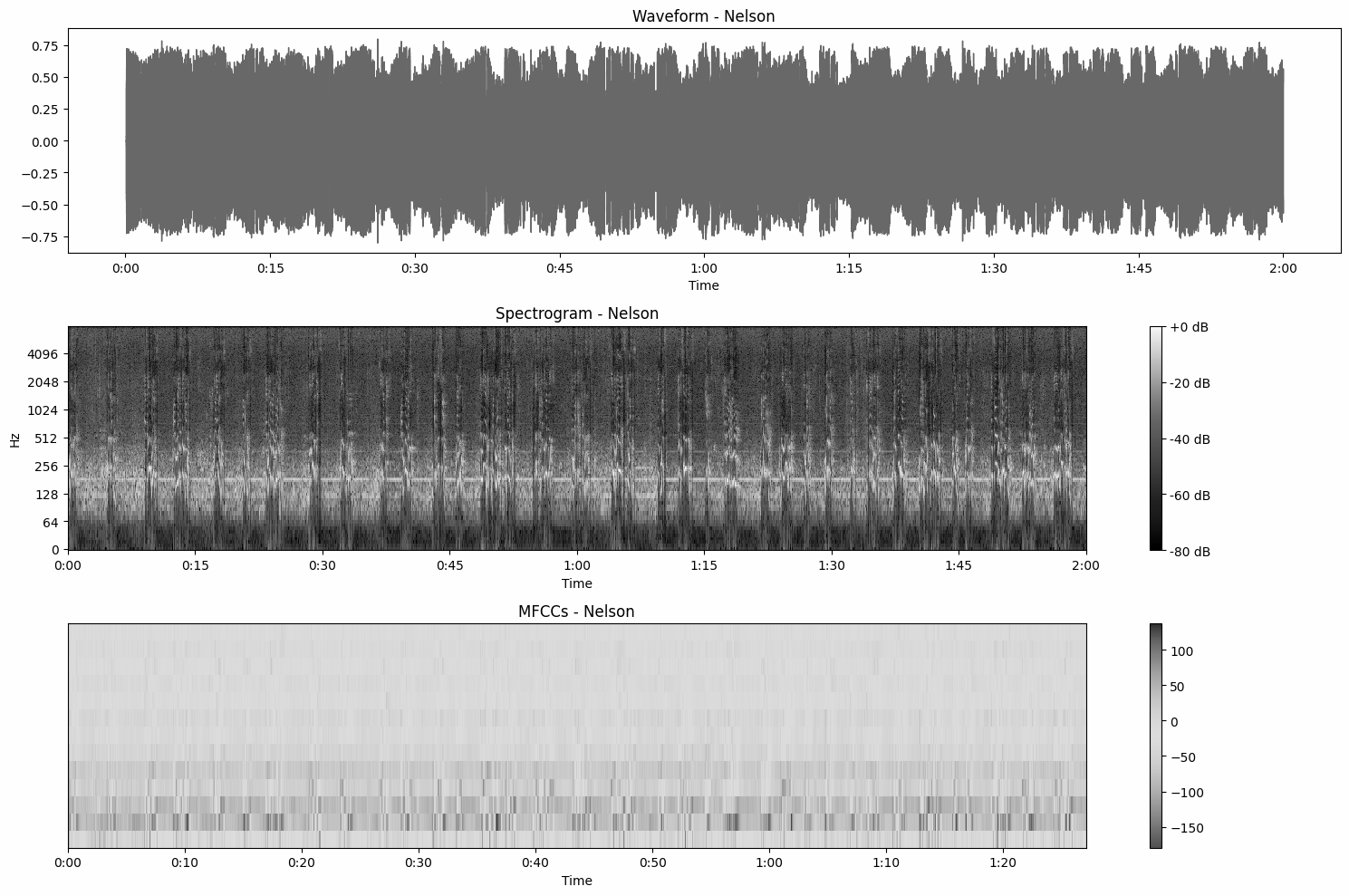}
  \caption{(a) Waveform, (b) spectrogram, and (c) MFCC graphs on voice identification model.}
  \label{fig:fig9}
\end{figure}

\section{Conclusion}
This research successfully demonstrates the feasibility of replicating the core functionalities of a system inspired by Smart Eye by developing modules for eye detection, facial expression recognition, and voice recognition. The eye detection model exhibits high accuracy and balanced precision and recall scores, indicating its reliability in determining eye state. Similarly, the facial expression identification model performs well in classifying diverse emotions. Finally, the voice recognition model demonstrates a strong ability to discriminate between speakers.

These findings highlight the potential of machine learning techniques for developing assistive technologies with applications in areas such as driver drowsiness detection, HCI, and healthcare monitoring. However, translating these individual modules into a cohesive, real-time system presents several additional challenges. Future research should focus on the integration of these components, optimising their interaction for real-time processing, and addressing issues of computational efficiency, particularly in resource-constrained environments.

This study establishes the modular efficacy of deep learning architectures for the three critical pillars of assistive perception: gaze, emotion, and voice. By validating that specialized, lightweight models (CNN and LSTM) can achieve high-performance metrics on discrete tasks, we provide a verified component library for future development. The immediate next step is the transition from discrete modular evaluation to a pipeline integration, focusing on latency optimization for real-time embedded deployment.

\section{Future Scope}
The potential for facial expression recognition extends far beyond a single project. Future research could focus on developing real-time applications for instant emotion analysis via webcams and smartphones \cite{littlewort2003real}. Further refinement of models to quantify the intensity of emotions would provide richer insights into nuanced emotional states. It's crucial to evaluate performance across diverse datasets representing various ethnicities and demographics. This will help mitigate biases and ensure the technology functions equitably. Transfer learning offers exciting possibilities to leverage pre-trained models and adapt architectures for improved facial expression recognition performance. To ensure accessibility, investigating model compression techniques will facilitate deployment on resource-constrained devices.

\section{Data Availability}
The datasets used in this study are publicly available. Any additional protocols or custom code used in this research were made available from the corresponding author upon reasonable request.

\begin{itemize}
\item The dataset used for Eye Detection, MRL Eye Dataset, was obtained via Kaggle and sourced from Media Research Lab (MRL) at the Department of Computer Science, VSB - Technical University of Ostrava. It can be accessed at kaggle.com.
\item The Facial Expression Recognition 2013 (FER2013) dataset, a benchmark dataset used for facial expression identification, is also publicly accessible at kaggle.com.
\item The audio data for voice recognition was drawn from a publicly available dataset of 1-second WAV audio files on Kaggle, accessible at kaggle.com.
\end{itemize}

\bibliographystyle{unsrt}  
\bibliography{references}  

\end{document}